\documentclass[journal,onecolumn,12pt]{IEEEtran}
\usepackage{amssymb}
\usepackage{amsmath}
\usepackage{graphicx}
\usepackage{tcolorbox}
\usepackage[ruled]{algorithm2e}
\usepackage{bm}
\usepackage{romannum}
\usepackage{xcolor}
\usepackage{enumitem}
\usepackage{hyperref}
\hypersetup{
    colorlinks=true,
    linkcolor=black,
    filecolor=black,      
    urlcolor=[rgb]{0.023529412,0.074509804,0.431372549},
}

\usepackage{lipsum}

\newcommand\blfootnote[1]{%
  \begingroup
  \renewcommand\thefootnote{}\footnote{#1}%
  \addtocounter{footnote}{-1}%
  \endgroup
}

\makeatletter
\newcommand{\subalign}[1]{%
  \vcenter{%
    \Let@ \restore@math@cr \default@tag
    \baselineskip\fontdimen10 \scriptfont\tw@
    \advance\baselineskip\fontdimen12 \scriptfont\tw@
    \lineskip\thr@@\fontdimen8 \scriptfont\thr@@
    \lineskiplimit\lineskip
    \ialign{\hfil$\m@th\scriptstyle##$&$\m@th\scriptstyle{}##$\hfil\crcr
      #1\crcr
    }%
  }%
}
\makeatother

\usepackage{siunitx}
\begin{document}
\title{From Learning to Meta-Learning: \\ Reduced Training Overhead and Complexity for Communication Systems}


\author{\large{Osvaldo Simeone\textsuperscript{*}\thanks{Codes for the results of this paper can be found at \url{https://github.com/kclip}.}, Sangwoo Park\textsuperscript{$\dagger$}, and Joonhyuk Kang\textsuperscript{$\dagger$}}\\ \vspace{10pt}  \small{\textsuperscript{*}King's
Communications, Learning \& Information Processing (KCLIP) Lab, \\ Department of Engineering, King's College London, London, United Kingdom}\\
\small{\textsuperscript{$\dagger$}School of Electrical Engineering, KAIST, Daejeon, South Korea}}

\pagenumbering{arabic}

\maketitle
\thispagestyle{plain}
\pagestyle{plain}

\vspace*{-1.6cm}
\begin{abstract}
Machine learning methods adapt the parameters of a model, constrained to lie in a given model class, by using a fixed learning procedure based on data or active observations. Adaptation is done on a per-task basis, and retraining is needed when the system configuration changes. The resulting inefficiency in terms of data and training time requirements can be mitigated, if domain knowledge is available, by selecting a suitable model class and learning procedure, collectively known as inductive bias. However, it is generally difficult to encode prior knowledge into an inductive bias, particularly with black-box model classes such as neural networks. Meta-learning provides a way to automatize the selection of an inductive bias. Meta-learning leverages data or active observations from tasks that are expected to be related to future, and a priori unknown, tasks of interest. With a meta-trained inductive bias, training of a machine learning model can be potentially carried out with reduced training data and/or time complexity. This paper provides a high-level introduction to meta-learning with applications to communication systems. 
\end{abstract}


\IEEEpeerreviewmaketitle

\section{Context and Motivation}
\label{sec:intro}

Adaptive algorithms have been a staple of communication systems for decades \cite{haykin2000adaptive, ibnkahla2000applications}. In Simon Haykin's words, ``a system is said to be adaptive when it is equipped with the ability
to respond positively to statistical variations of the environment in which it operates" \cite{haykin2000adaptive}. Adaptivity hence describes a data-driven behavior, whereby the operation of a system is modified so as to optimize a given performance criterion in response to an unknown or changing environment on the basis of past observations. Through such a process, adaptivity can also be useful as a tool to automatically discover effective solutions when techniques optimized ``by hand'' are too complex to implement. 

In today's preferred terminology, adaptive algorithms carry out Machine Learning (ML) or, using more high-flown language, instantiate ``Artificial Intelligence'' for the given narrow tasks under study. The resurgence of interest in the use of ML methods in communication networks is to be attributed, on the one hand, to the increasing complexity of such systems, which makes optimal engineered solutions often impractical; and, on the other hand, to the wide availability of data and of advanced software and hardware solutions for ML, especially for the training of neural networks. Through adaptation, ML can help tackle model deficiencies (the lack of well established models for the given environment), as well as algorithmic deficiencies (the lack of efficient engineered solutions) \cite{simeone2018very}.  

\begin{figure}[t!]
    \centering
    \vspace*{-1cm}
    \includegraphics[width=0.45\columnwidth]{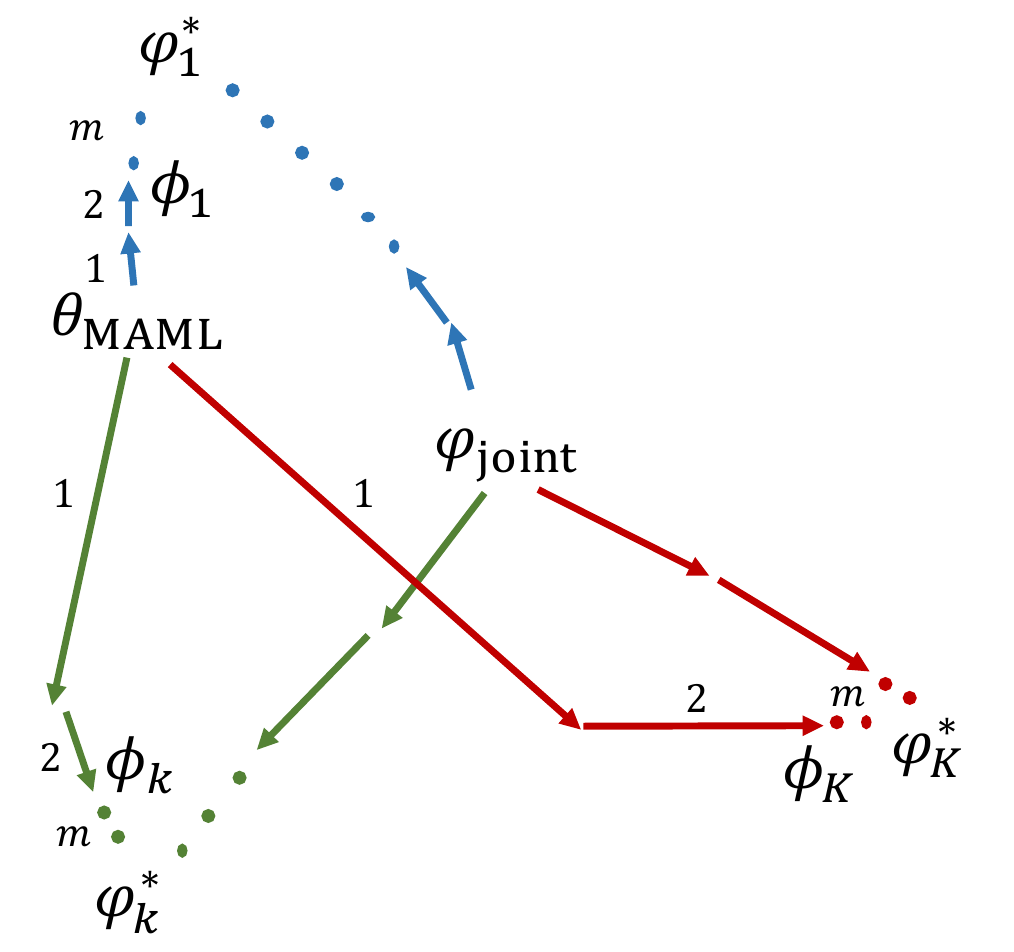}
    \caption{When faced with multiple system configurations, a standard joint learning solution attempts to find a single solution $\varphi_\text{joint}$ that is as ``close'' as possible to the optimal trained models $\varphi^{*}_k$ for each configuration $k$. Model Agnostic Meta-Learning (MAML), a specific example of meta-learning algorithms \cite{finn2017model}, aims instead at finding a common intialization point $\theta_\text{MAML}$ that allows for a quick \emph{adaptation} towards each optimal trained model $\varphi^{*}_k$. Adaptation is based on a small number $m$ of training steps, and correspondingly few data points, for each task (from \cite{fallah2019convergence}).}
    \vspace*{-0.5 cm}
    \label{fig:fig_1}
\end{figure}

 Conventional ML methods start by assuming an \emph{inductive bias} and are predicated on the availability of \emph{sufficient data and training time} for the given system configuration of interest. The inductive bias encompasses a \emph{model class}, such as neural networks with a given architecture; and a \emph{learning procedure}, typically defined by a learning criterion, such as the average squared loss, and by an optimization algorithm, such as Stochastic Gradient Descent (SGD) with given hyperparameters. A \emph{system configuration} here refers to a given stationary behavior of the system, e.g., by a channel realization for a communication link or by a traffic distribution for the routers in the core network. 
 

In terms of \emph{data requirements}, \emph{supervised learning} trains based on a data set of input-output examples for the given system configuration, e.g., pairs of pilot symbols and received signals for the training of a receiver \cite{ibnkahla2000applications}; \emph{unsupervised learning} only requires either inputs or outputs, which may be possibly produced by the learner itself, e.g., inputs for the training of a transmission link in the presence of a channel model \cite{o2017introduction}; and \emph{reinforcement learning} is based on a direct interaction of the learner with the environment, or an emulated version thereof, e.g., for the transmission over an unknown channel model \cite{aoudia2019model}. 

As for \emph{training time requirements}, given the stochastic nature of standard training procedures, iterations are needed both to process a sufficiently large number of data points and to explore the space of solutions within the selected model \cite{shalev2014understanding, simeone2018brief} (see also the more modern viewpoint detailed in \cite{belkin2019reconciling}). 

The discussed potential inefficiency of conventional ML in terms of data and time requirements can be mitigated if domain knowledge is available. Domain knowledge can, in fact, guide the selection of a suitable inductive bias, e.g., a model class with a specific structure \cite{zappone2019wireless}. However, it is generally difficult to encode prior knowledge into an inductive bias, particularly with black-box model classes such as neural networks. Meta-learning, or learning to learn \cite{thrun1998lifelong}, provides a way to automatize the selection of an inductive bias by leveraging data or active observations from tasks that are known to be related to future, a priori unknown, system configurations of interest. This paper provides a brief introduction to meta-learning with applications to communication systems. We start by providing a high-level discussion in the next section, and we then discuss mathematical formulations in Section III and Section IV. Section V provides a brief review of recent use cases for communication systems.


\section{From learning to meta-learning}

Protocols and algorithms for communication networks are expected to operate in a variety of system configurations. For example, encoders and decoders should be able to adapt their operation to the current channel realization, and network data analytics tools should modify their operation to changing traffic statistics. \emph{Conventional learning} would train a separate model for each configuration, incurring the data and training time costs described in Section I. 

As a first potential solution, one could attempt to train a single model within the selected model class that is able to perform as well as possible on \emph{all} configurations (see Fig. 1). The downside of this approach, which we refer to as \emph{joint training}, is that there may not be a single trained model that obtains a satisfactory performance on all tasks. For instance, an encoder and decoder pair trained by minimizing an average performance loss over a given fading channel distribution can only perform as well as a non-coherent modulation and coding scheme \cite{o2017introduction}.

Meta-learning operates at a higher abstraction level, using data from multiple configurations -- \emph{tasks} in meta-learning parlance -- not to identify a single model compatible with an a priori inductive bias, but to infer an inductive bias that enables efficient learning on each system configuration of interest \cite{thrun1998lifelong,baxter1998theoretical}. The inferred inductive bias may refer to the selection of a model class, e.g., through a feature extractor \cite{vinyals2016matching}; or of a learning procedure, e.g., through its learning rate \cite{maclaurin2015gradient}, optimization update rule \cite{bengio1990learning, wichrowska2017learned, li2017meta, flennerhag2019meta}, or initialization \cite{finn2017model} (see Fig. 1). 

To understand the difference in abstraction level between learning and meta-learning, consider the problem of \emph{few-shot classification} \cite{vinyals2016matching}. The goal of few-shot classification is to infer a learning procedure that can train an effective classifier based on a limited number of training examples for each class. For instance, we may wish to quickly train a classifier of images of different objects -- say, birds vs bicycles -- based on the observation of a few images of either class. To this end, meta-learning algorithms are provided with data from multiple, related, few-shot classification tasks, e.g., cats vs dogs, apples vs pears, and so on. This \emph{meta-training} data is used to infer a learning procedure that can quickly train a classifier on the corresponding \emph{meta-training tasks}, and not to train a single model to classify well across all such tasks.

An illustration of the operation of a specific, popular, meta-training method known as Model Agnostic Meta-Learning (MAML) \cite{finn2017model} is provided in Fig. 1. Unlike standard joint learning, which aims at finding a single solution $\varphi_\text{joint}$ that is as ``close'' as possible to the optimal trained models $\varphi^{*}_k$ for each task/configuration $k$, the goal of MAML is to find a common intialization point $\theta_\text{MAML}$ that allows for a quick \emph{adaptation} towards each optimal trained model $\varphi^{*}_k$. This will be further discussed in Section IV.


\section{Conventional and Joint Learning}
\label{sec:background}
In this section, we provide a mathematical formulation for conventional learning and joint learning, while MAML is described as an example of meta-learning in the next section. Throughout the following two sections, we focus on supervised and unsupervised learning with deterministic models. While the same concepts apply more broadly, probabilistic models and reinforcement learning require more complex notations and derivations.

\subsection{Conventional Learning}
\label{subsec:conv_learning}
\begin{figure}[t!]
    \centering
    \includegraphics[width=0.35\columnwidth]{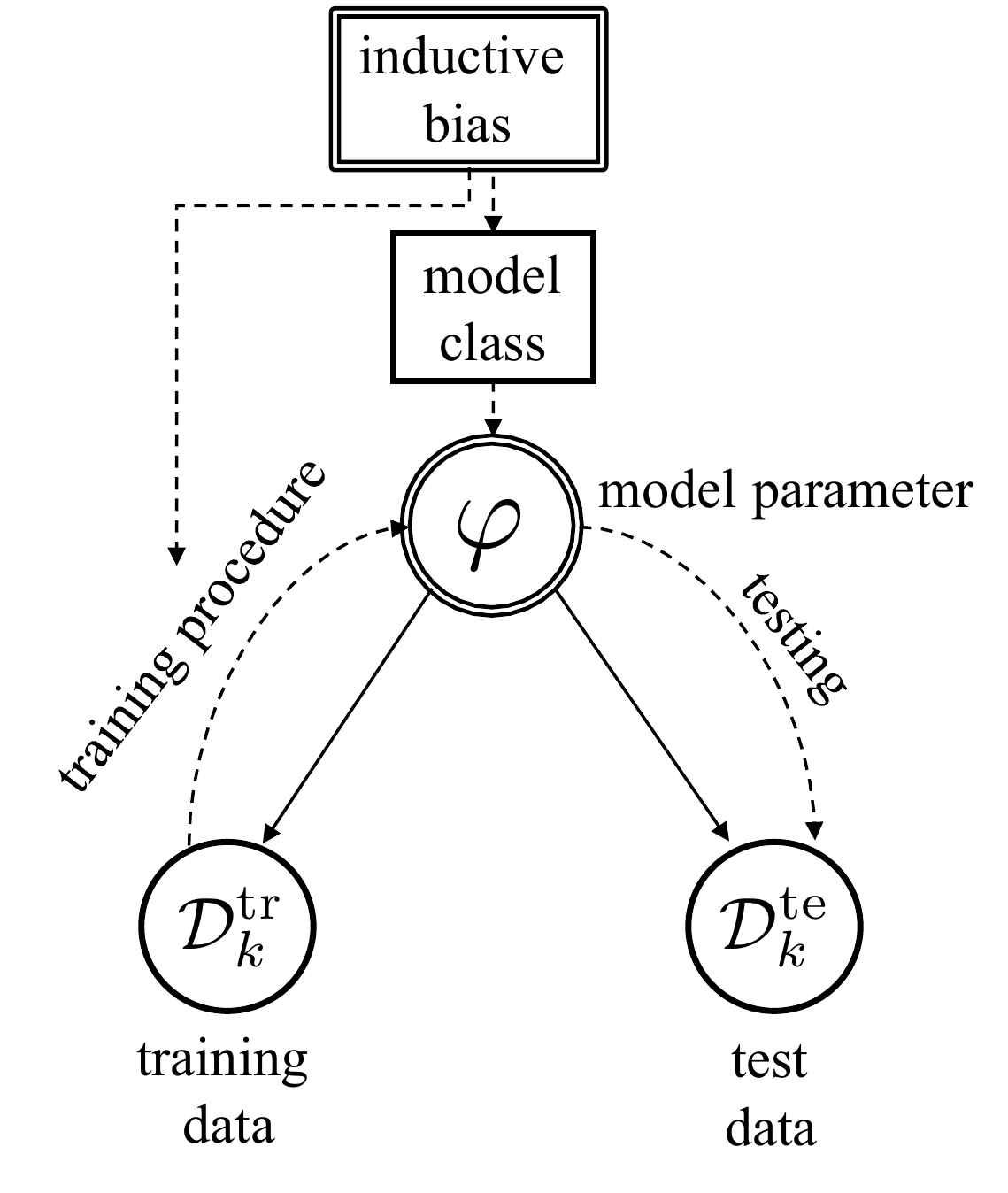}
    \caption{Conventional learning: A predefined inductive bias determines model class and training procedure. The model parameter $\varphi$ is trained based on training data $\mathcal{D}_k^\text{tr}$ separately for any given system configuration, or task, $k$, and tested on a separate set $\mathcal{D}_k^\text{te}$ of test data. \vspace{-0.5cm}}
    \label{fig:conv}
\end{figure}
As illustrated in Fig. \ref{fig:conv}, conventional learning focuses on a single task, or system configuration, $k$. Having fixed an inductive bias in the form of a model class and a training procedure, conventional learning applies the learning procedure on the training set $\mathcal{D}_k^\text{tr}$ for the given task in order to train a task-specific model parameterized by vector $\varphi$. The trained parameter vector $\varphi^{*}_k$ is then tested on a separate test set $\mathcal{D}_k^\text{te}$ for the same task. 

Mathematically, the typical goal of conventional learning is to minimize the out-of-sample, or population, or generalization, loss
\begin{align} \label{eq:exp-loss}
L_k(\varphi) = \mathbb{E}_{x\sim P_k}[\ell(x,\varphi)],
\end{align}
where the expectation is taken over the true distribution $P_k$ of data point $x$ for task $k$, and function $\ell(x,\varphi)$ measures the loss for data point $x$ when using the model characterized by parameter vector $\varphi$ within the given model class. In a learning problem, the distribution $P_k$ is usually assumed to be unknown to the learner, which only has access to a training data set $\mathcal{D}_k^\text{tr} = \{ x \sim P_k \}$ of examples sampled from $P_k$. Therefore, in lieu of the population loss in \eqref{eq:exp-loss}, most learning procedures minimize (some function of) the \emph{training loss}
\begin{align} \label{eq:emp-loss}
L_{\mathcal{D}_k^\text{tr}}(\varphi) = \sum_{x \in \mathcal{D}_k^\text{tr}}\ell(x,\varphi),
\end{align}
which is an empirical approximation of the population loss \eqref{eq:exp-loss} \cite{shalev2014understanding, simeone2018brief}. 

Most learning procedures tackle the minimization of \eqref{eq:emp-loss} via some form of Stochastic Gradient Descent (SGD) \begin{align} \label{eq:SGD}
\varphi \leftarrow \varphi - \eta \nabla_\varphi \ell(x,\varphi),
\end{align}
where example $x$ is drawn at random from the training set $\mathcal{D}_k^\text{tr}$ and and $\eta$  is a learning step size (see \cite{goodfellow2016deep,simeone2018brief} for extensions and variants).

Once training is completed, the performance is assessed by estimating the population loss \eqref{eq:exp-loss} through an empirical average over a separate, test, data set $\mathcal{D}_k^\text{te}$ of examples also drawn from the task-specific distribution $P_k$, which is computed as $L_{\mathcal{D}_k^\text{te}}(\varphi) = \sum_{x \in \mathcal{D}_k^\text{te}}\ell(x,\varphi)$. 

\subsection{Joint Learning}
\label{subsec:joint_learning}

\begin{figure}[t!]
    \centering
    \includegraphics[width=0.35\columnwidth]{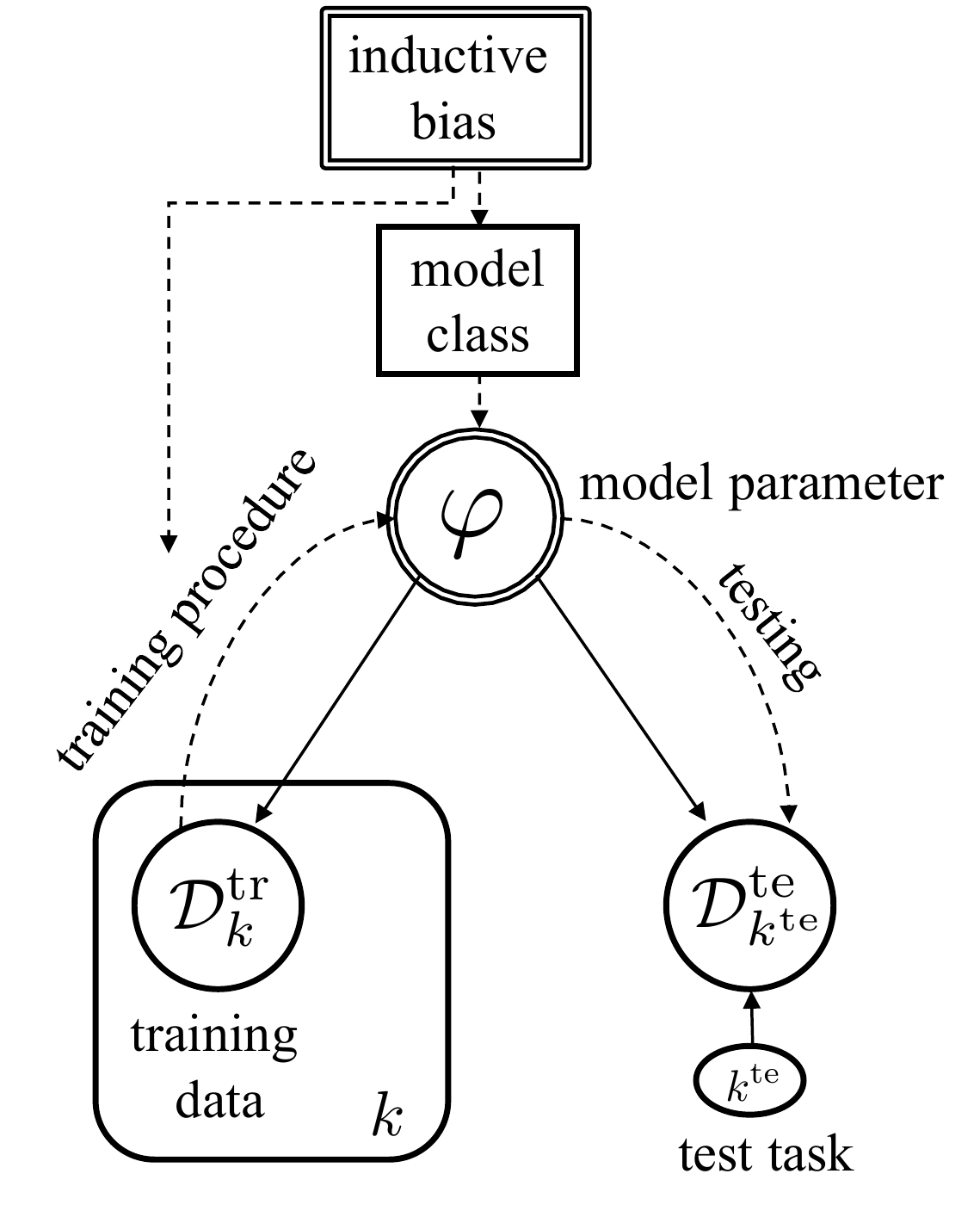}
    \caption{Joint learning: A predefined inductive bias determines model class and training procedure. The model parameter $\varphi$ is trained based on training data $\mathcal{D}_k^\text{tr}$ from a set of related tasks $k$ (shown in the figure with the tile notation \cite{koller2009probabilistic}). Performance is tested on a test data set $\mathcal{D}^\text{te}_{k^\text{te}}$ for a randomly selected new task $k^\text{te}$.\vspace{-0.5cm}}
    \label{fig:joint}
\end{figure}

Assume now that we are faced with a distribution of tasks of interest as illustrated in Fig. \ref{fig:joint}. As an example, to be further elaborated on in Section V, we may wish to train a decoder to operate over a link in which the channel, defining the task, is well modelled by a Rayleigh fading distribution. For a fixed inductive bias, while conventional learning would retrain a ML model from scratch for every task, joint training aims at finding a single model, through parameter vector $\varphi$, that optimizes a performance loss on average over all tasks. 

Mathematically, the goal is to minimize the population loss 
\begin{align} \label{eq:joint-exp-loss}
L(\varphi) = \mathbb{E}_{k\sim Q} [L_k(\varphi)],
\end{align}
where $L_k(\varphi)$ is the population loss for the $k$th task in \eqref{eq:exp-loss} and the expectation is taken over task distribution $Q$. In words, joint training carries out conventional training on the mixture data distribution $P=\sum_k Q(k)P_k$. Accordingly, define as $K$ the number of tasks for which a training set $\mathcal{D}_k^\text{tr}$ is available. Joint training aims at minimizing (a function of) the  \emph{joint training loss} 
\begin{align} \label{eq:joint-emp-loss}
L_{\mathcal{D}}(\varphi) = \sum_{k=1}^{K} L_{\mathcal{D}_k^\text{tr}}(\varphi),
\end{align}
where $L_{\mathcal{D}_k^\text{tr}}(\varphi)$ is the task-specific training loss in \eqref{eq:emp-loss}. 

As illustrated in Fig.~1, if the optimal parameters $\varphi^{*}_k$ for different tasks are significantly different, joint training may converge to a solution that does not generalize well on a randomly sampled test task $k^\text{te}\sim Q$.



\section{Meta-Learning}
\label{sec:meta-learning}
\begin{figure}[t!]
    \centering
    \includegraphics[width=0.73\columnwidth]{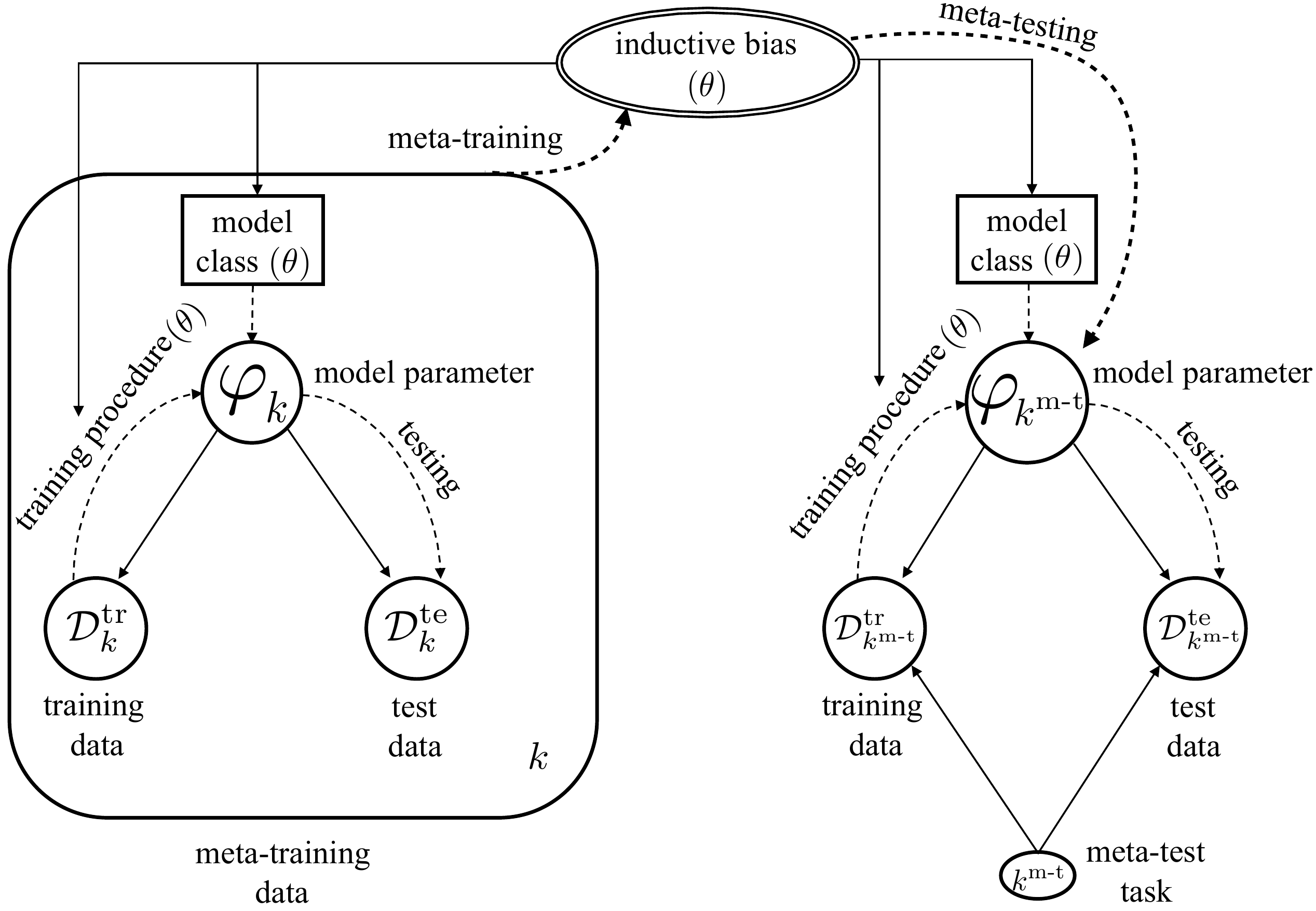}
    \caption{Meta-learning: Based on data from multiple ``meta-training'' tasks $k$ (shown in the figure with the tile notation \cite{koller2009probabilistic}), a inductive bias, parameterized by a vector $\theta$, is determined and applied to train a randomly selected ``meta-test'' task $k^\text{m-t}$. For the latter, conventional learning is carried out using the ``meta-trained'' inductive bias $\theta$ on the basis of a (typically small) training data set $\mathcal{D}_{k^\text{m-t}}^\text{tr}$. Performance is tested on a test data set $\mathcal{D}_{k^\text{m-t}}^\text{te}$ for the meta-test task $k^\text{m-t}$.\vspace{-0.5cm}}
    \label{fig:meta}
\end{figure}

As illustrated in Fig. \ref{fig:meta}, meta-learning assumes that, once deployed, an ML algorithm has access to, typically few, training examples to adapt to the new task of interest. For example, once deployed, an ML-based decoder may have access to pilot symbols for the current channel realization. Meta-learning uses data from previously observed tasks in order to infer an inductive bias that allows for a ``fast'' adaptation on a new task using the task-specific training examples. 

To use the most common terminology, a meta-learning algorithm is given data for a number $K$ of meta-training tasks, with tasks drawn from some distribution $Q$ (see previous section). For each meta-training task $k$, the meta-learner has available a training set $\mathcal{D}_k^\text{tr}$ and a test data set  $\mathcal{D}_k^\text{te}$, both with examples drawn from the corresponding distribution $P_k$. The inductive bias, described by a ``shared'' parameter $\theta$, is inferred based on the overall meta-training data $\mathcal{D}_k^\text{tr}$ and $\mathcal{D}_k^\text{te}$ for all $k=1,...,K$. Once meta-training is complete, the performance of the inductive bias $\theta$ is tested by drawing a new, meta-test, task $k^\text{m-t}$ from $Q$ and corresponding data sets $\mathcal{D}_{k^\text{m-t}}^\text{tr}$ and $\mathcal{D}_{k^\text{m-t}}^\text{te}$. The inductive bias is applied to the training set $\mathcal{D}_{k^\text{m-t}}^\text{tr}$ and tested on data set $\mathcal{D}_{k^\text{m-t}}^\text{te}$.

Meta-learning algorithms differ mostly in terms of the type of inductive bias that is optimized for and on the application, e.g., few-shot classification or reinforcement learning (see Section \ref{sec:intro}). A fairly general formulation of the problem can be given by formalizing the set-up in Fig. \ref{fig:meta} through a probabilistic graphical model, with the ``context'' variables $\varphi$ treated as latent random variables. The meta-learning problem can then be formulated either as the frequentist problem of estimating the shared parameter $\theta$ via Expectation Maximization (EM) \cite{grant2018recasting} or as a fully Bayesian inference problem \cite{yoon2018bayesian}. In practice, the complexity of an exact implementation of these general approaches calls for simplifications that have been widely investigated in recent years (see, e.g., \cite{finn2017model, nichol2018first, nguyen2019uncertainty, chen2019modular}).

Before introducing one of the most popular such solutions, namely MAML, it is worth mentioning that a poorly selected inductive inductive bias can impair adaptation as compared to a generic one. It is therefore of theoretical and practical interest to characterize the amount of meta-training data that guarantees an effective ``meta-generalization'' on a new task. This problem has been addressed via Probably Approximately Correct (PAC) theory in \cite{baxter1998theoretical}, and, more recently, within a PAC Bayes framework in \cite{pentina2014pac, amit2017meta}. We also refer to \cite{yin2019meta} for discussion on meta-generalization and meta-overfitting.




Mathematically, MAML aims to infer a shared parameter vector $\theta$ to initialize $m$ SGD updates \eqref{eq:SGD} for any task $k\sim Q$. Such updates produce an adapted parameter $\varphi_k$ for any task $k\sim Q$. For instance for $m=1$, we have the adaptation step\begin{align} \label{eq:maml-local}
\varphi_k = \theta - \eta \nabla_\theta L_{\mathcal{D}_k^\text{tr}}(\theta),
\end{align}where we have used the full training loss for clarity of notation, although SGD is typically used. The goal is to ensure that the \emph{population meta-training loss} 
\begin{align} \label{eq:maml}
L^{\text{MAML}}(\theta) = \mathbb{E}_{k \sim Q} [\mathbb{E}_{x \sim P_k}[\ell(x,\varphi_k)]]
\end{align}
obtained \emph{after adaptation}, that is, on the adapted parameter vector $\varphi_k$ in (\ref{eq:maml-local}) is minimized.  Approximating the expectations in \eqref{eq:maml} with empirical averages, MAML estimates the criterion (\ref{eq:maml}) via the meta-training empirical loss
\begin{align} \label{eq:maml-compute}
L^\text{MAML}_\mathcal{D}(\theta) = \sum_{k=1}^K L_{\mathcal{D}_k^\text{te}}(\varphi_k),
\end{align}with $K$ randomly sampled meta-training tasks through distribution $Q$. Note that, while the local adaptation updates \eqref{eq:maml-local} use the training part of the data for a meta-training task $k$, the generalization loss in (\ref{eq:maml-compute}) is estimated using the test part of the same data. Optimization of the meta-learning criterion is carried out via SGD with respect to the shared parameters $\theta$. For $m=1$, this is given as
\begin{align} \label{eq:maml-meta}\nabla_\theta L^\text{MAML}_\mathcal{D}(\theta) = \sum_{k=1}^K (\mathbf{J}_\theta \varphi_k) \nabla_{\varphi_k} L_{\mathcal{D}_k^\text{te}}(\varphi_k) \\
= \sum_{k=1}^K(I-\eta \nabla_\theta^2 L_{\mathcal{D}^\text{tr}_k}(\theta) )\nabla_{\varphi_k}L_{\mathcal{D}_k^\text{te}}(\varphi_k),
\end{align} where $\eta$ is a learning rate and $\mathbf{J}_\theta$ represents the Jacobian operator (i.e., the matrix of derivatives of the argument) with respect to $\theta$. An important observation is that, since the derivatives need to ``flow through'' the local update, MAML is a second-order scheme, requiring second derivatives of the loss functions.

\section{Use Cases}
\label{sec:use-cases}

In this section, we discuss two use cases for meta-learning by implementing MAML for the supervised learning of demodulator or a decoder \cite{park2019learning,jiang2019mind} and for the end-to-end unsupervised learning of encoder and decoder of a communication link \cite{park2019meta}. These use cases will highlight the potential of meta-learning as a means to reduce both data and training time requirements (see Section I). For other applications, see \cite{yangdeep}.

\begin{figure}[t!]
    \centering
    \includegraphics[width=0.55\columnwidth]{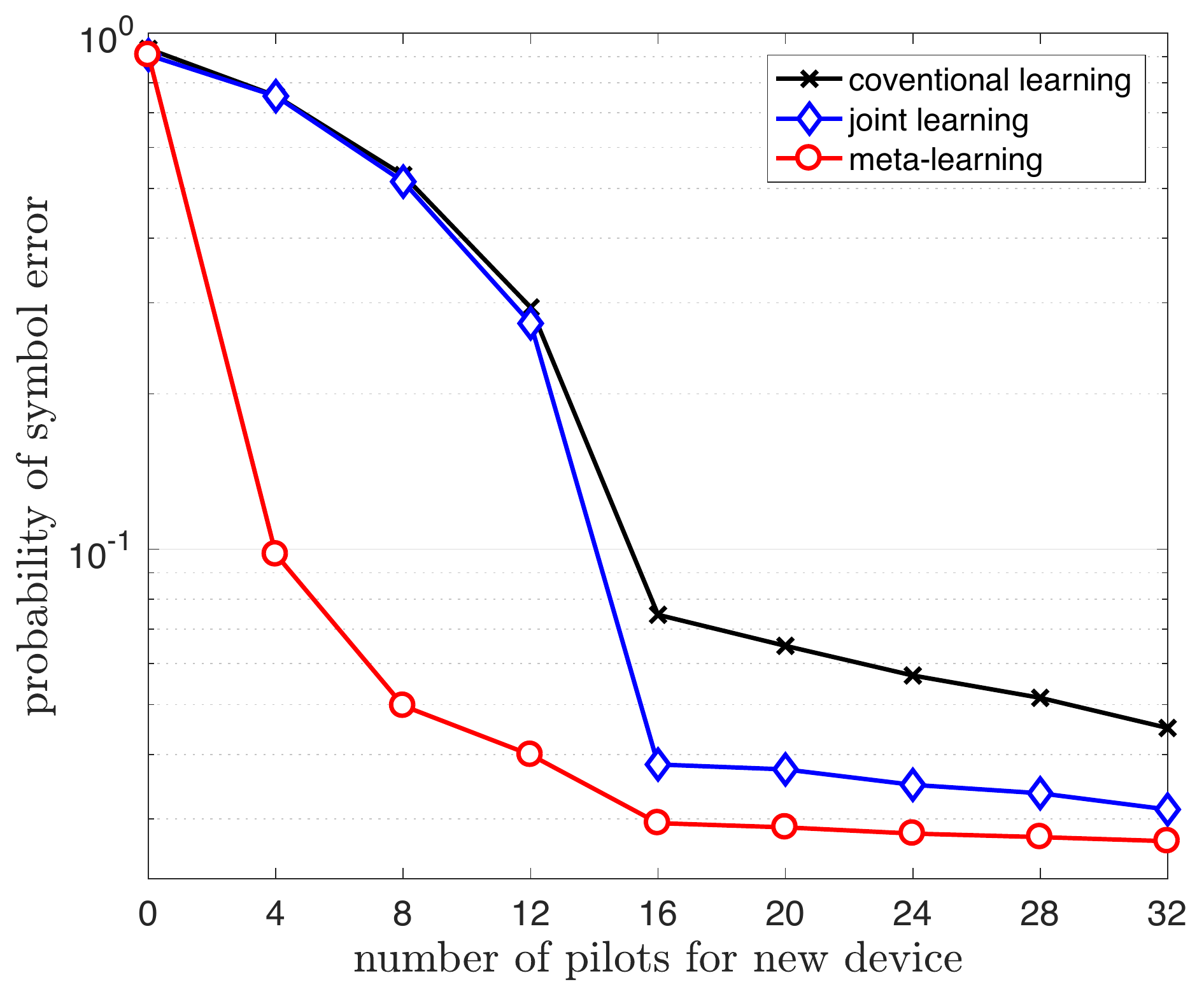}
    \caption{Probability of symbol error with respect to number of pilots for a new device assuming 16-QAM, Rayleigh fading, and transmitters' non-idealities ($m=1$, see \cite{park2019learning} for details).
    \label{fig:supervised}}
\end{figure}

\emph{Few-Pilot Supervised Learning for Demodulation.} Consider an Internet-of-Things (IoT)-like scenario in which each device transmits sporadically using short packets with few pilot symbols over a fading channel characterized by hard-to-model non-linearities and fading. The goal is to design a receiver that can adapt its operation based on such few pilots via meta-learning. With meta-learning, pilots received from other devices can be used to infer an inductive bias (an initialization with MAML) that allows a quick adaptation on any new (meta-test) device. Fig.~\ref{fig:supervised}, fully described in \cite{park2019learning}, shows the probability of symbol error with respect to number of pilots for a new device. Meta-learning can clearly outperform conventional training and joint training, even when we allow for adaptation based on the received pilots. 

\begin{figure}[t!]
    \centering
    \includegraphics[width=0.55\columnwidth]{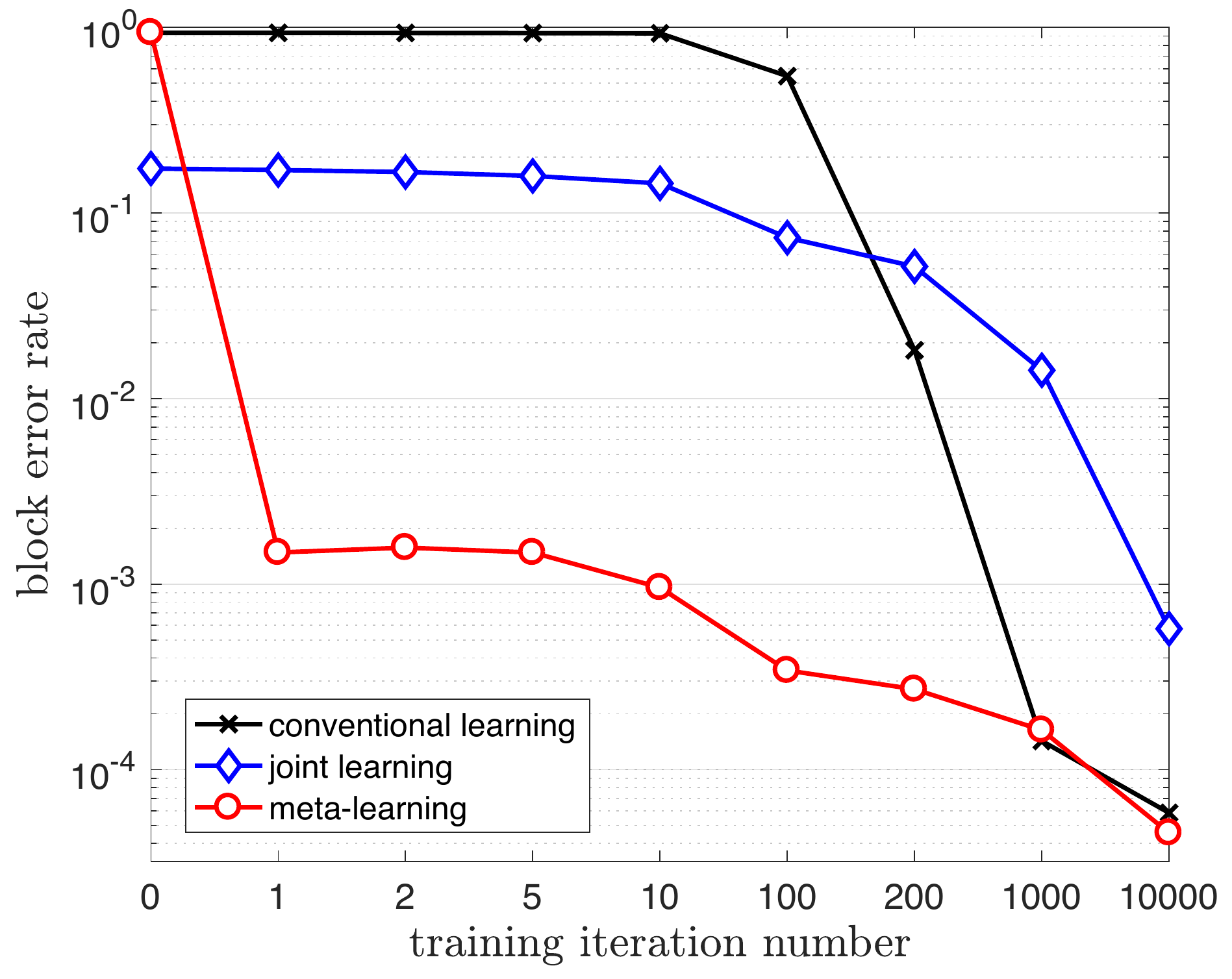}
    \caption{Block error rate over iteration number for the unsupervised training on a new channel assuming Rayleigh block fading channel model with three taps ($m=1$, see \cite{park2019meta} for details)}
    \vspace*{-0.5cm}
    \label{fig:unsupervised}
\end{figure}

\emph{Fast Unsupervised Learning for Transmission and Reception.} We now consider the problem of training an autoencoder to provide encoding and decoding for an end-to-end fading communication link with Gaussian noise  \cite{o2017introduction}. In the presence of a channel model, training can be carried out in an unsupervised manner, whereby fading and noise samples are generated as needed. Using meta-learning based on data from previously observed channel configurations,  Fig.~\ref{fig:unsupervised}, adapted from \cite{park2019meta}, shows that the block error rate can be made to decrease very quickly with a limited number of training iterations. 

\small
\section{Acknowledgments}
\label{sec:ack}

The work of O. Simeone was supported by the European Research Council (ERC) under the European Union's Horizon 2020 research and innovation programme (grant agreement No. 725731). The work of S. Park and J. Kang was supported by the National Research Foundation of Korea (NRF) grant funded by the Korea government (MSIT) (No. 2017R1A2B2012698).

\bibliographystyle{IEEEtran}
\bibliography{ref}

\end{document}